\def\year{2020}\relax
\documentclass[letterpaper]{article} 
\usepackage{aaai20}  
\usepackage{times}  
\usepackage{helvet} 
\usepackage{courier}  
\usepackage[hyphens]{url}  
\usepackage{graphicx} 
\usepackage{graphicx}  
\usepackage{soul}
\usepackage{url}
\usepackage{hyperref}
\usepackage[small]{caption}
\usepackage{amsmath}
\usepackage{amsfonts}       
\usepackage{booktabs}
\usepackage{latexsym}

\usepackage{algorithm}
\usepackage{algorithmic}
\usepackage{natbib}
\title{Explanation vs Attention: A Two-Player Game to Obtain Attention for VQA}
\author{Badri N. Patro, \Large \textbf{Anupriy,} \Large \textbf{Vinay P. Namboodiri}\\ 
Indian Institute of Technology, Kanpur \\ 
\{badri,anupriy,vinaypn\}@iitk.ac.in \\ 
}
 
\begin{document}
\maketitle

\begin{abstract}
In this paper, we aim to obtain improved attention for a visual question answering (VQA) task. It is challenging to provide supervision for attention. An observation we make is that visual explanations as obtained through class activation mappings (specifically Grad-CAM) that are meant to explain the performance of various networks could form a means of supervision. However, as the distributions of attention maps and that of Grad-CAMs differ, it would not be suitable to directly use these as a form of supervision. Rather, we propose the use of a discriminator that aims to distinguish samples of visual explanation and attention maps. The use of adversarial training of the attention regions as a two-player game between attention and explanation serves to bring the distributions of attention maps and visual explanations closer. Significantly, we observe that providing such a means of supervision also results in attention maps that are more closely related to human attention resulting in a substantial improvement over baseline stacked attention network (SAN) models. It also results in a good improvement in rank correlation metric on the VQA task. This method can also be combined with recent MCB based methods and results in consistent improvement. We also provide comparisons with other means for learning distributions such as based on Correlation Alignment (Coral), Maximum Mean Discrepancy (MMD) and Mean Square Error (MSE) losses and observe that the adversarial loss outperforms the other forms of learning the attention maps. Visualization of the results also confirms our hypothesis that attention maps improve using this form of supervision.
\end{abstract}
\vspace{-1.1em}

\section{Introduction}

 
 
  When asked a question based on an image, a human invariably focuses on the part of the image that aids in answering the question. This is a commonly known fact in cognitive science. An extreme example that depicts perceptual blindness was demonstrated by \citep{simons1999gorillas}, where two groups of participants are passing balls. When asked to count the balls, viewers ignore a gorilla in the video as it is not pertinent to the task of counting. However, the deep networks that solve semantic tasks such as visual question answering do not have such attentive mechanisms. The fact that the existing deep networks do not attend to the areas that humans do was shown by the work of \citep{Das_EMNLP2016}. While there have been some works that aim to improve the attended regions, it is challenging as obtaining supervision for attention is tedious and may not always be possible for all the semantic tasks that we would like to use deep networks. In this paper, we propose a simple method to obtain self-supervision to guide attention.


The main idea is that given the task of solving visual question answering (VQA), there exist methods based on obtaining visual explanations such as Grad-CAM \citep{selvaraju2017grad} that obtain class activation mappings from gradients that allow us to understand the areas that a network focuses while solving the task for the correct class label. As during training, class labels are available for the VQA task; it is easy to obtain such supervision. Using this, it is possible to obtain surrogate supervision for supervising attention. One can obtain the visual explanation using the ground-truth label for a deep network that solves the visual question answering task. As the network is provided the actual label, the corresponding activation maps do aid in solving the task. Therefore, we hypothesize that this supervision can aid in obtaining better attention maps, and this is evident from the results that we obtain.

The next challenge is to consider how the surrogate supervision obtained from Grad-CAM can be used to obtain better attention regions. Directly using these as supervision is not optimal as the distributions for the visual explanation differs from that of the attention maps as the attention maps are also supervised by the task loss. We show that just using the mean-square error loss for the two maps is sub-optimal. In this paper, we show that a very simple way of using a two-player game between a discriminator that tries to discriminate between Grad-CAM results \& attention maps and a generator that generates attention maps serves to provide substantially improved attention maps. We show that this method performs much better and also provides state of the art results in terms of attention maps that correlates well with human attention maps. To summarize, through this paper, we provide the following contributions :

\begin{itemize}
    \item We propose a means for obtaining surrogate supervision for obtaining better attention maps based on the visual explanation in the form of Grad-CAM results. This method performs better as compared to other forms of surrogate supervision such as using RISE \citep{petsiuk_BMVC2018rise}.
    \item We show that this surrogate supervision can be best used through a variant of adversarial learning to obtain attention maps that correlate well with the visual explanation. Further, we observe that this performs better as against other means of supervision, such as MMD \citep{tzeng_Arxiv2014deep} or Coral \citep{sun_ECCV2016deep} losses.
    \item We provide various comparisons and results to show that we obtain better attention maps that correlate well with human attention maps and outperform other techniques for VQA. Further, we show that obtaining better attention maps also aids in obtaining better accuracies while solving for VQA. A detailed empirical analysis for the same is provided.
\end{itemize}    
   

\subsection{Motivation}
In VQA, given an image \& a query, the attention model aims to learn the regions in an image pertinent to the answer. ~\citep{Das_EMNLP2016} has proposed Human Attention (HAT) dataset for VQA task where human annotators have annotated the regions attended in the image to mark the answer based on the question. The regions pointed by humans for answering the visual question are more accurate as compared to those obtained by other techniques. This can be concluded through an experiment on HAT dataset where we replace human attention with attention obtained using stacked attention network with one stack. We observe that the prediction accuracy increases with ground truth human attention map for the stacked attention network ~\citep{Yang_CVPR2016}.  We believe that human attention cannot be directly used as supervision, as there are not enough examples of human attention (58K/215K). Further, such a method would not generalize well to novel tasks. However, we are motivated by this result and have therefore developed in this paper a self-supervision based method to improve attention.  We formulate a game between Attention vs. Explanation using adversarial mechanism. Through this game, we observe that we obtain improved attention regions, which lead to improved prediction and therefore, also results in better regions obtained through visual explanation as shown in the figure-~\ref{fig:result_1_B}. Thus, improving attention using Grad-CAM results in an improvement in Grad-CAM too. To ensure whether our approach is prudent, we evaluate whether using grad-CAM as self-supervision is beneficial. We do this by an experiment that replaces attention mask with Grad-CAM mask, and we observed that the classification accuracy of the VQA (SAN) model increases by ~4\% on the validation set. This provides a strong intuition to consider using Grad-CAM as self-supervision for the attention module.  

 \vspace{-0.5em}
\section{Related work}\label{sec:lit_surv}
Visual question answering (VQA) was first proposed by ~\citep{Malinowski_NIPS2014}. Subsequently, ~\citep{Geman_PNAS2015} proposed a "Visual Turing test" where a binary question is generated from a given test image. This is in contrast to modern approaches in which the model is trying to answer free-form open-ended questions. A seminal contribution here has been standardizing the dataset used for Visual Question Answering ~\citep{VQA}. The methods for VQA can be categorized into joint embedding approaches and attention based approaches. Joint embedding based approaches have been proposed by ~\citep{VQA,Ren_NIPS2015,Goyal_CVPR2017,Noh_CVPR2016} where visual features are combined with question features to predict the answer. Attention based approaches are the other category of methods for solving VQA. It comprises of image based, question based and some that are both image and question based attention. \citep{Shih_CVPR2016} has proposed an image based attention approach, the aim is to use the question in order to focus attention over specific regions in an image.~\citep{Yang_CVPR2016} has proposed a method to repeatedly obtain attention by using stacked attention over an image based on the question. Our work uses this as one of the baselines. \citep{Li_NIPS2016} has proposed a region based attention model over images. Similarly, ~\citep{Zhu_CVPR2016,Xu_ECCV2016,bao_EMNLP2018deriving} have proposed interesting method for question based attention. A work that explores joint image and question includes that is based on hierarchical co-attention is \citep{Lu_NIPS2016}. There has been interesting work by ~\citep{Fukui_arXiv2016,Kim_ICLR2017,kim_NIPS2018bilinear,Patro_CVPR2018dvqa} that advocates multimodal pooling and obtains close to state of the art results in VQA.  

The task of VQA is well studied in the vision and language community, but it has been relatively less explored for providing explanation ~\citep{selvaraju2017grad, Goyal_CVPR2017} for answer prediction. We start with image captioning ~\citep{Socher_TACL2014, Vinyals_CVPR2015, Karpathy_CVPR2015, Xu_ICML2015, Fang_CVPR2015, Chen_CVPR2015, Johnson_CVPR2016, Yan_ECCV2016} to provide a basic explanation for an image. The next level of challenging task is to provide an explanation for the visual question answering system. The attention-based model provides some short of basic explanation for VQA. This is observed that models \citep{Das_EMNLP2016} are not looking at the same regions as humans are looking. So we need to improved model attention and as well as an explanation for our answer prediction. Recently, Uncertainty based explanation method \citep{Patro_2019_ICCV} is proposed to improve the attention mask for VQA. ~\citep{jain2019attention} has proposed a method to evaluate how attention weights can provide a correct explanation in language prediction task. There are very interesting methods to provide visual explanations such as Grad-CAM \citep{selvaraju2017grad}, RISE \citep{petsiuk_BMVC2018rise}, U-CAM \citep{Patro_2019_ICCV}. In contrast to the above-mentioned approaches, we focus on improving image-based attention using an adversarial game between Grad-CAM and attention mask and show that it correlates better with human attention. Our approach allows the use of visual explanation as a means for obtaining surrogate supervision for attention.

 \begin{figure*}[ht]
	\centering
	\includegraphics[width=0.8\textwidth]{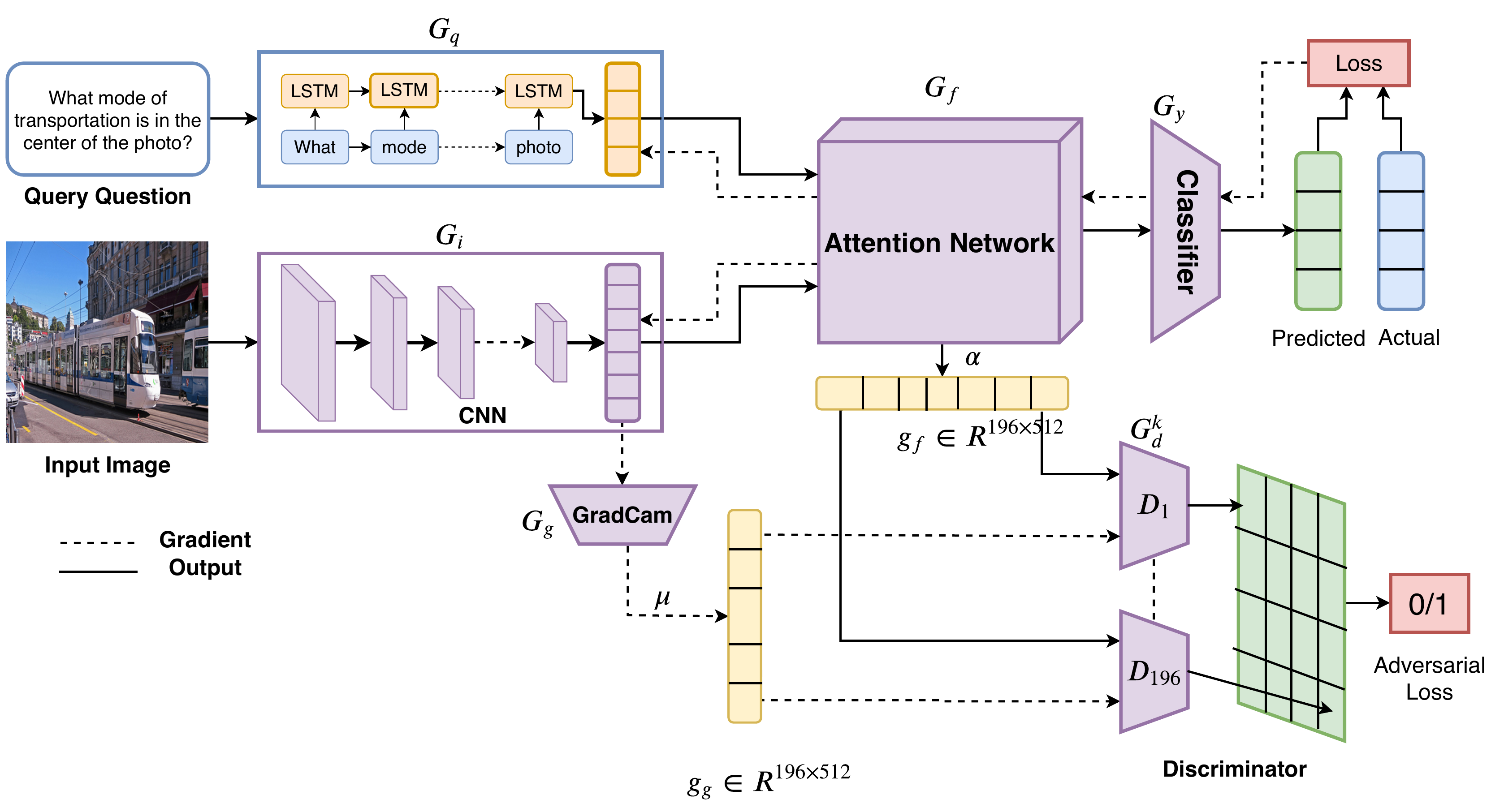}
	\vspace{-1em}
	\caption{Illustration of model PAAN and its attention mask. Image feature and question feature are obtained using CNN and LSTM respectively. Attention mask is then obtained using these features and classification of the answer is done based on the attended feature. We  have  improved  the  attention  mask  with  the visual explanation approaches based on Grad-CAM}
	\label{fig:main}
	\vspace{-1em}
\end{figure*}
\section{Method}
The main focus in our approach for solving visual question answering (VQA) is to use supervision obtained from visual explanation methods such as Grad-CAM to improve attention. As mentioned earlier, using Grad-CAM as attention shows improved performance in comparison to just using attention alone. Therefore, we believe that Grad-CAM, or any other visual explanation method can be used in this setting. Further, by learning both the visual explanation and attention jointly in an adversarial setting, we observe improvements in both as shown empirically. 

The key differences in our architecture as compared to an existing VQA architecture is the use of visual explanation and attention blocks in an adversarial setting. This is illustrated in figure~\ref{fig:main}. The other aspects of VQA are retained as is. In particular, we adopt a classification based approach for solving VQA where an image embedding is combined with the question embedding to solve for the answer. This is done using a softmax function in a multiple choice setting:
$\hat{A}=\underset{A \in \Omega}{argmax}P(A|I,Q, \theta)$,
where $\Omega$ is a set of all possible answers, and $\theta$ represents the parameters in the network. 
\subsection{Our Approach}
The three main components of our approach, as illustrated in figure \ref{fig:main} are 1) Attention representation,  2)Explanation representation, 3) Adversarial Game.  The details of our method are provided in the following sub-sections:

\subsubsection{Attention Representation}
Initially, we obtain an embedding $g_i$ for an image $X_i$ using a convolution neural network (CNN). Similarly, we obtain a question feature embedding $g_q$ for the query question $X_Q$ using an LSTM network. These are input to an attention network that combines the image and question embeddings using a weighted softmax function and produces a weighted output attention vector $g_f$. There are various ways of modeling the attention network. In this paper, we have evaluated the network proposed in SAN~\citep{Yang_CVPR2016} and MCB~\citep{Fukui_arXiv2016}. 

\subsubsection{Explanation Representation}
One of the ways for understanding a result obtained by a deep network is to use visualization strategies. One such strategy that has gained acceptance in the community is based on Grad-CAM~\citep{selvaraju2017grad}. Grad-CAM uses the gradient information of the last convolutional layer to visualize the contribution of each pixel in predicting the results. Note that Grad-CAM uses ground-truth class information and finds the gradient of the score for a class $c$ in a convolution layer. It averages the gradient values to find the averaged  $\mu$ values for each of the channels of the layer. We follow this approach, and further details are provided in ~\citep{selvaraju2017grad}. We have also evaluated with another such approach termed as RISE \citep{petsiuk_BMVC2018rise}. We observed better results using Grad-CAM.





\subsubsection{Adversarial Game}
A zero-sum adversarial game between two players (P1, P2) is used with one set of players being a Generator network and the other being a discriminator network. They choose a decision from their respective decision sets $\mathcal{K}_1$ and $\mathcal{K}_2$. In our case, the attention network is the generator network, and the `real' distribution is the output of Grad-CAM network. We term the resultant network as `Adversarial Attention Network' (AAN). A game objective $ \mathcal{L}(G,D) :\mathcal{K}_1 \times \mathcal{K}_2 \in \mathcal{R}$, sets the utilities of the players. Concretely, by choosing a proper strategy $(G, D) \in \mathcal{K}_1 \times \mathcal{K}_2$ the utility of P1 is $ -\mathcal{L}(G, D)$, while the utility of P2 is $\mathcal{L}(G, D)$. The goal of either P1/P2 is to maximize their worst case utilities; thus,
\begin{equation}
 \begin{split}
& \min_{G \in \mathcal{K}_1 } \max_{D \in \mathcal{K}_2} L(G,D) \quad \text{(Goal of P1)},\\  & \max_{D \in \mathcal{K}_2} \min_{G \in \mathcal{K}_1 } L(G,D)  \quad \text{(Goal of P2)}
 \end{split}
\end{equation}
The above formulation raises the question of whether there exists a solution$ (G^*, D^*)$  to which both players may jointly converge. The solution to this question is to obtain a Nash equilibrium where the Discriminator is unable to distinguish the generations of the Generator network from the `real' distribution {\it{i.e.}} $[\max_{D \in \mathcal{K}_2} L(G^*,D) = \min_{G \in \mathcal{K}_1 }L(G,D^*)]$.


Since pure equilibrium does not always exist \cite{nash1950equilibrium}, there exists an approximate solution for this issue as a Mixed Nash Equilibrium,i.e. 
\begin{equation}
\max_{D \in \mathcal{K}_2} \mathbb{E}_{G \sim D_1} L(G,D) = \min_{G \in \mathcal{K}_1 } \mathbb{E}_{D \sim D_2} L(G,D) 
\end{equation}
Where $D_1$ is the distribution over $K_1$, and $D_2$ is the distribution over $K_2$. In zero-sum adversarial game, the sum of the generator’s loss and the discriminator’s loss is always zero, i.e. the generator’s loss is: $\mathcal{L}^G = - \mathcal{L}^D$. The solution for a zero-sum game is called a minimax solution, where the goal is to minimize the maximum loss. We can summarize the entire game by stating that the loss function is $L^G$ (which is the discriminator’s payoff), so that the minimax objective is
\begin{equation*}
\begin{split}
\min_{G} \max_{D} L_{1}(G,D)&  = E_{g_{g_i} \sim G_g(x_i)} [\log D(g_{g_i}/x_i)] + \\ &E_{g_{f_i} \sim G_f(x_i)} [\log (1- D(G(g_{f_i}/x_i)))]
\end{split}
\end{equation*}
For simplicity, we remove subscript $i$. Here $g_{g}$ is the output of Grad-cam network $G_g$ for a sample, $x_i$ and $g_{f}$ is the output of the attention network. The discriminator wants to maximize the objective (i.e., its payoff) such that $D(g_g/x)$ is close to 1 and $D(G(g_{f}/x))$ is close to zero. The generator wants to minimize the objective (i.e., its loss) so that D(G(z)) is close to 1. Specifically, the discriminator is a set of CNN layers followed by a linear layer that uses a binary cross entropy loss function. In case we have access to ground-truth attention obtained from humans, we can directly use this in our framework. Here, we assume that we do not have access to such ground-truth as it is challenging to obtain this and is being used only for evaluation.

The final cost function for the network combines the loss obtained through an adversarial loss for the attention network along with the cross-entropy loss while solving for VQA. The final cost function used for obtaining the parameters $\theta_f$ of the attention network, $\theta_y$ of the classification network, and $\theta_d$ for the discriminator is as follows:
\begin{equation}
\label{e4}
C(\theta_f,\theta_y,\theta_d)=\frac{1}{n}\sum_{j=1}^{n}{ (L^j_c(\theta_{f},\theta_y)} + \eta L^j(\theta_{f},\theta_d))
\end{equation}
Where n is the number of examples, and $\eta=10$ is the hyper-parameter, fine-tuned using validation set and $L_c$ is standard cross entropy loss.  We train the model with this cost function till it converges so that the parameters $(\hat{\theta}_f,\hat{\theta}_y,\hat{\theta}_d)$ deliver a saddle point function.
	\begin{equation}
    \begin{split}
        & (\hat{\theta}_{f},\hat{\theta}_y)= \arg\max_{\theta_f,\theta_y}(C(\theta_f,\theta_y,\hat{\theta}_d))\\
        & (\hat{\theta}_{d})= \arg\min_{\theta_d}(C(\hat{\theta}_f,\hat{\theta}_y,\theta_d))\\
    \end{split}
\end{equation}
      

\textit{Pixel-wise Adversarial Attention Network (PAAN)}\label{PAAN}:
A variation of the adversarial attention network is to obtain a local pixel-wise discriminator for obtaining an improved attention network. The idea of pixel-wise discriminators has been studied for generative adversarial networks (GANs) and is termed patch-GAN. We show here, that doing pixel-wise (with multiple channels per pixel) attention network results in an improved attention network. We term this network a Pixel-wise Adversarial Attention Network (PAAN). Though this network uses more local discrimination, it does not increase the parameters of the network as compared to AAN. The effect of local discrimination results in improved attention as well as explanation. The algorithm for training the same is provided in Algorithm ~\ref{alg:PAAN}. The resultant min-max loss function is obtained as follows:
\begin{equation*}
\label{e6}
\begin{split}
\min_{G} \max_{D^k} L^k_{1}(G,D^k) = E_{g_{g_i} \sim G_g(x_i)} [\log D^k(g_{g_i}/x_i)] +\\ E_{g_{f_i} \sim G_f(x_i)} [\log (1- D^k(G(g_{g_i}/x_i)))]
\end{split}
\end{equation*}

Finally, the actual cost function for training the pixel-wise discriminator, attention network and Grad-CAM is given by:
\begin{equation*}
\label{e7}
C(\theta_f,\theta_y,\theta_d^k|_{k=1}^K)=\frac{1}{n}\sum_{j=1}^{n}{ \big( L^j_c(\theta_{f},\theta_y) + \eta \sum_{k=1}^{K}{  L^{j,k}}(\theta_{f},\theta_d^k)}\big)
\end{equation*}
The main problem we face is the model convergence issue where the model parameter oscillates and does not converge using gradient descent in a minimax game. To handle convergence issue, we add  JS-divergence \cite{fuglede2004jensen} to the cost function, which penalizes a poor generated mask badly as compared to a good one, which is different from  KL-divergence \citep{kullback1951information}. The second issue faced is ``vanishing gradient", when discriminator is successful (which can well distinguish between generated and discriminator sample), then the generator gradient vanishes and learns nothing. To handle this issue, we add Pearson-$\chi^2$ divergence \citep{mao2017least} to the GAN cost function.
\vspace{-0.7em}
\begin{algorithm}[htb]
  \caption{Training PAAN}
  \label{alg:PAAN}
\begin{algorithmic}
\scriptsize
  \STATE {\bfseries Input:}  Image $X_I$, Question $X_Q$
  \STATE {\bfseries Output:}  Answer $X_A$
  \REPEAT
    \STATE  \text{Attention features  $ G_f(G_i(X_I),G_q(X_Q))\gets g_a$}
    \STATE  \text{Classification score  $ G_y(g_a)\gets  \hat{y}$}
    \STATE  \textit{Answer cross entropy $L_y \gets$ loss$(\hat{y},y)$}
    \STATE \textit{Compute Gradient,$L_f= \frac{\partial L_y}{ \partial  \theta_y}$,$L_i= \frac{\partial L_f}{ \partial  \theta_f}$}
    \STATE  \textit{update $\theta_c$ $\gets$ $\theta_c$ - $ \frac{\partial L_c}{ \partial  \theta_c}$} 
    \STATE  \text{Explanation features $ f_t(\theta_f,X_t) \gets X_t$}
    \REPEAT 
          \STATE \textit{Sample fake mini batch(Attention): $\alpha_1 \dots \alpha_{196}$}
          \STATE \textit{Sample real mini batch(Gradient): $\mu_1 \dots \mu_{196}$}
            \STATE  \text{Discriminator: $ D^r_k(\mu_k)\gets d^r_k,D^f_k(\alpha_k)\gets d^f_k$}            
           \STATE \textit{Update the discriminator  by ascending its stochastic gradient\\ $ \nabla_{\theta_d}\frac{1}{m}\sum_{i=1}^{m} [ \log D(\mu_k) + \log(1-D(\alpha_k))]$ }
           \UNTIL{$k=1:K$}
        \STATE \textit{Sample fake mini batch(Attention): $\alpha_1 \dots \alpha_{196}$}
         \STATE \textit{Update the Generator  by descending its stochastic gradient: $ \nabla_{\theta_g}\frac{1}{m}\sum_{i=1}^{m} \log(1-D(\alpha))$ }
    \UNTIL{Number of Iteration}
    
\end{algorithmic}
\end{algorithm}

  \begin{table*}[t]
     \begin{minipage}{0.5\textwidth}
 	  \begin{tabular}{|l|c|c|}
    \hline
    \textbf{Model} & \textbf{RC($\uparrow$)} & \textbf{EMD($\downarrow$)}\\
    \hline
    SAN \citep{Das_EMNLP2016}& 0.2432 & 0.4013\\
    CoAtt-W \citep{Lu_NIPS2016}	& {0.246 } & --\\		
	CoAtt-P \citep{Lu_NIPS2016}	& {0.256 }&--\\
	CoAtt-Q \citep{Lu_NIPS2016}	& {0.264 }&--\\\hline
	MMD\_RISE& 0.2591 & 0.3992 \\
    Coral\_RISE & 0.2609 & 0.3978 \\
    MSE\_RISE& 0.2622 & 0.3921 \\
    AAN\_RISE& 0.2683 & 0.3900 \\
    PAAN\_RISE& 0.2754 & 0.3894 \\\hline
    MMD (ours) & 0.2573 & 0.3895 \\
    Coral (ours) & 0.2563 & 0.3851 \\
    MSE (ours)& 0.2681 & 0.3814 \\
    AAN (ours)& 0.2896 & 0.3721 \\
    \textbf{PAAN} (ours) & \textbf{0.3071} & \textbf{0.3701} \\\hline
    PAAN\_Ran\_07& 0.1213 & 0.6700 \\
    PAAN\_Ran\_20& 0.1746 & 0.5872 \\\hline
	Human \cite{Das_EMNLP2016} & { 0.623 } &--\\ \hline
  \end{tabular}
  \vspace{-0.3cm}
  \caption{Attention mask comparison for SOTA \& Ablation Methods}
  \label{tab_rank_correlation}
        \end{minipage}
        \hspace{-0.8cm}
        \begin{minipage}{0.6\textwidth}
            \centering
		\begin{tabular}{|l|c|c|c|c|  } \hline
			\textbf{Models} & \textbf{All} & \textbf{Yes/No} & \textbf{Num} & \textbf{Oth}  \\ \hline 
			Baseline-ATT & { 56.7 }& {78.9 }& { 35.2}& { 36.4} \\
			MMD\_SAN\_RISE   &{ 56.9 }& {79.1 }& { 35.8 }& { 38.1} \\ 
		 	Coral\_SAN\_RISE & { 57.4 }& {79.8 }& { 36.0 }& { 39.6} \\ 
			MSE\_SAN\_RISE & 58.2 & 80.1 & 36.4& 40.2 \\
			AAN\_SAN\_RISE & 59.3 & 80.4 & 36.9 &  42.5 \\ 
            PAAN\_SAN\_RISE & 60.1 & 80.8 & 37.3 & 44.2 \\ \hline
			MMD\_SAN\_GCAM   &{ 58.9 }& {80.3 }& { 37.0 }& { 43.7} \\ 
		 	Coral\_SAN\_GCAM & { 59.4 }& {80.8 }& { 36.5 }& { 45.1} \\ 
			MSE\_SAN\_GCAM & 60.8 & 80.0 & 36.8& 47.1 \\
			AAN\_SAN\_GCAM & 62.3 & 80.4 & 37.2 &  49.8 \\ 
            PAAN\_SAN\_GCAM & 63.6 & 81.1 & 36.9 & 50.9 \\ \hline
            AAN\_MCB\_GCAM & 66.4 & 84.6 & 37.8 &  54.7 \\ 
            PAAN\_MCB\_GCAM & \textbf{67.1} & \textbf{85.0} & \textbf{38.4} & \textbf{55.9} \\\hline
            PAAN\_SAN\_Ran\_07 & 55.2 & 77.2 & 35.1& 36.2 \\
			PAAN\_SAN\_Ran\_20 & 57.3 & 78.7 & 35.6 &  39.7 \\\hline
		\end{tabular}
		\vspace{-0.7em}
		\caption{Ablation analysis for Open-Ended VQA1.0 accuracy on test-dev}
		\label{abl_VQA1_accuracy}
        \end{minipage}
        \vspace{-1.8em}
    \end{table*}

\vspace{-1.2em}
\subsection{Variations of Proposed Method}
\label{subsec:variants}
While we advocate the use of Adversarial explanation method for improving the attention mask, we also evaluate several other explanation methods for this architecture. Our intuition is that, if we learn an attention mask that minimizes the distance between attention probability distribution and the gradient class activation map, then we are more easily able to train our VQA classifier module to provide correct answer. To minimize these distances we have used various methods.

\textbf{Maximum Mean Discrepancy (MMD) Net}: In this variant, we minimize this distance using MMD~\citep{tzeng_Arxiv2014deep} based standard distribution distance metric. We have computed this distance with respect to a representation $\psi(.)$. In our case, we obtain representation feature $\psi(\alpha)$ for attention \& $\psi(\mu)$ for Grad-CAM map.

  \textbf{CORAL Net}: In this variant, we minimize distance between second-order statistics (co-variances) of attention and Grad-CAM mask using CORAL loss~\citep{sun_ECCV2016deep} based standard distribution distance metric. Here, both $(\mu)$ and $(\alpha)$ are the d-dimensional deep layer activation feature for attention and Grad-CAM maps.  We have computed feature co-variance matrix of attention feature and Grad-cam feature represented by $C(\alpha)$ and $C(\mu)$ respectively.
 
 We trained  our variants MMD and CORAL  directly without adversarial loss to bring Grad-CAM based pseudo distribution close to attention distribution. Finally we replace MMD and CORAL with adversarial loss.

\vspace{-0.7em}
\section{Experiment}
We evaluate the proposed method i.e. PAAN in a number of ways which includes both quantitative analysis and qualitative analysis. Quantitative analysis includes ablation analysis with other variants that we tried using metrics such as Rank correlation (RC) score \citep{Das_EMNLP2016}, Earth Mover Distance (EMD) \citep{arjovsky_STAT2017wasserstein}, and VQA accuracy etc. as shown in table ~\ref{tab_rank_correlation} and ~\ref{abl_VQA1_accuracy} respectable. We also compare our proposed method with various state of the art models, as provided in table \ref{VQA1_accuracy} and ~\ref{VQA2_accuracy}. Qualitative analysis includes visualization of improvement in attention maps for some images as we move from our base model to the PAAN model. We also provide visualization of Grad-CAM maps for all the models. 

\vspace{-0.5em}
\subsection{Ablation analysis on model parameter}\label{AblationQuant}
We provide comparisons of our proposed model PAAN and other variants along with base model using various metrics in the table \ref{tab_rank_correlation} and table \ref{abl_VQA1_accuracy}. Rank correlation and EMD score are calculated for each model against human attention map \citep{Das_EMNLP2016}. Each model's generated attention map is used for this purpose. The rank correlation has an increasing trend. Increase in rank correlation indicates the dependency of the attention maps that are compared. As rank correlation increases, attention map generated from the model and human attention map becomes more dependent. In other words, higher rank correlation shows similarity between the maps. EMD also improves for PAAN. To verify our intuition, that we can learn better attention mask by minimising the distance between attention mask and explanation mask, we start with MMD and observe that both rank correlation and answer accuracy increase by 1.42 and 1.2 \% from baseline respectively. Also, we observe that with CORAL and MSE based distance minimisation technique, both RC and EMD improves as shown in the table-~\ref{tab_rank_correlation}. Instead of the predefined distance minimisation technique, we adapt an adversarial learning method. The proposed AAN method improves attention globally with respect to Grad-CAM. AAN improves 3.9\% in-terms of RC and 9.5\% on VQA accuracy.  Finally,our proposed PAAN, which considers local pixel-wise discriminator improves 6.4\% in RC and 10.4\% in VQA accuracy as mentioned in the table \ref{tab_rank_correlation} and table \ref{abl_VQA1_accuracy}. Since, human attention map \citep{Das_EMNLP2016} is only available for VQA-v1 dataset, for VQA accuracy we perform ablation for VQA-v1 only. However, we provide state of the art results for both datasets (VQA-v1 and VQA-v2). 

  \begin{table*}[t]
        \begin{minipage}{0.5\textwidth}
            \centering
 		\begin{tabular}{|l|c|c|c|c|} \hline
			\textbf{Models} & \textbf{All} & \textbf{Y/N} & \textbf{Num} & \textbf{Oth}  \\ \hline 
			 Baseline-ATT & { 56.7 }& {78.9 }& { 35.2}& { 36.4} \\
			DPPnet \citep{Noh_CVPR2016}    & { 57.2 }& {80.7 }& { 37.2 }& { 41.7} \\
		 	SMem (\citeauthor{Xu_ECCV2016}) &{ 58.0 }& {80.9 }& { 37.3 }& { 43.1} \\ 
			SAN \citep{Yang_CVPR2016}      & { 58.7 }& {79.3 }& { 36.6 }& { 46.1 }\\
			DMN \citep{Xiong_arXiv2016} & {60.3 }& { 80.5 }& { 36.8 }& { 48.3 } \\ 
			QRU(2) \citep{Li_NIPS2016}  & { 60.7 }& {82.3 }& { 37.0 }& { 47.7 }\\				
			HieCoAtt \citep{Lu_NIPS2016} & {61.8}& { 79.7 }& { 38.9 }& { 51.7 } \\
			MCB \citep{Fukui_arXiv2016} & {64.2}& { 82.2 }& { 37.7 }& { 54.8 } \\
			MLB \citep{Kim_ICLR2017} & {65.0}& { 84.0} & { 37.9 }& { 54.7 }\\
			DVQA \citep{Patro_CVPR2018dvqa} &65.4& 83.8& 38.1& 55.2
			\\ \hline
			AAN + SAN (ours)& 62.3 & 80.4 & 37.2 &  49.8 \\ 
            PAAN + SAN(ours)& 63.6 &81.1 & 36.9 & 50.9 \\ 
		    AAN + MCB (ours)& 66.4 & 84.6 & 37.8 &  54.7 \\ 
            PAAN + MCB (ours)& \textbf{67.1} & \textbf{85.0} & \textbf{38.4} & \textbf{55.9} \\ \hline
		\end{tabular}
		\vspace{-0.7em}
		\caption{SOTA: Open-Ended VQA1.0 accuracy on test}
		\label{VQA1_accuracy}
        \end{minipage}
        \hspace{0.2cm}
        \begin{minipage}{0.5\textwidth}
            \centering
             \begin{tabular}{|l|c|c|c|c|  } \hline
			\textbf{Models} & \textbf{All} & \textbf{Y/N} & \textbf{Num} & \textbf{Oth}  \\ \hline 
			SAN-2 \citep{Yang_CVPR2016}     & {54.9 }& {74.1 }& {35.5 }& {44.5 }\\ 
			MCB  \citep{Fukui_arXiv2016} & {64.0}& { 78.8 }& {38.3 }& {53.3 } \\
			DVQA \citep{Patro_CVPR2018dvqa}&65.9& 82.4& 43.2& 56.8\\
			MUTAN \citep{Ben_ICCV2017} & {66.0}& { 82.8 }& {44.5}& { 56.5 } \\
			MLB \citep{Kim_ICLR2017} & {66.3}& {83.6} & {44.9}& {56.3 } \\
			DA-NTN \citep{bai_ECCV2018deep} & {67.5}& {84.3} & {47.1}& {57.9} \\
			Counter \citep{zhang_ICLR2018learning}& 68.0&83.1&51.6&58.9\\
			
			GCA \citep{Patro_2019_ICCV}& 69.2 & {85.4} & 50.1 & 59.4 \\
			BAN \citep{kim_NIPS2018bilinear} & {69.5}& {85.3}&\textbf{50.9}&{60.2}\\
			BU \citep{anderson_CVPR2018bottom}& \textbf{70.34}& \textbf{86.6}&{48.64}&\textbf{61.15}\\\hline
			AAN + SAN (ours) & 60.1 &76.4 & 35.2 & 51.8 \\ 
            PAAN + SAN (ours)& 61.3 &78.0 & 38.6 & 52.9 \\ 
			AAN + MCB (ours) & 67.6 & 84.8 & 47.5 &  57.7 \\ 
            PAAN +MCB (ours)& \textbf{68.4} & \textbf{85.1} & \textbf{48.4} & \textbf{59.1} \\\hline
		\end{tabular}%
		\vspace{-0.7em}
		\caption{SOTA: Open-Ended VQA2.0 accuracy on test}
		\label{VQA2_accuracy}
        \end{minipage}
        \vspace{-1.8em}
    \end{table*}

\subsection{Ablation on Explanation: Why do we select Grad-CAM?}\label{AblationExp}
 \vspace{-0.5em}
While calculating Grad-CAM one uses the ``true” class labels in obtaining activation maps. When observing attention, one just infers these for a sample without using the ground-truth label. At test time, Grad-CAM results cannot be used as true class labels would not be available. By using Grad-CAM as supervision, the aim is to obtain dense supervision for the attention module that will guide the attention methods as against the sparse rewards that are available based on the correct classification prediction. To validate this we conduct an experiment with another kind of visual explanation, i.e., RISE \citep{petsiuk_BMVC2018rise} in a similar way to  Grad-CAM\citep{selvaraju2017grad}. In RISE, we use the true label to obtain RISE based activation maps, instead of Grad-CAM, that corresponds to the true prediction. This, as surrogate supervision, is observed by us to provide better results as compared to using only attention without supervision. We evaluate the rank correlation of the attention mask for RISE supervision and observe that it is much lower than Grad-CAM supervision, as shown in table-\ref{tab_rank_correlation}.  This method results in an improvement of 3.22\% in terms of rank correlation over the baseline SAN \citep{Das_EMNLP2016} method while we obtain an improvement of 6.39\% using Grad-CAM supervision. Similarly, we observe that the Earth Mover Distance of RISE based model is higher than the Grad-CAM based model. We believe that the suggested framework can always be improved by any other surrogate supervision technique that can be developed.
\vspace{-1.2em}
\begin{figure}[htb]
	\centering
	\includegraphics[width=0.47\textwidth,height=2.85cm]{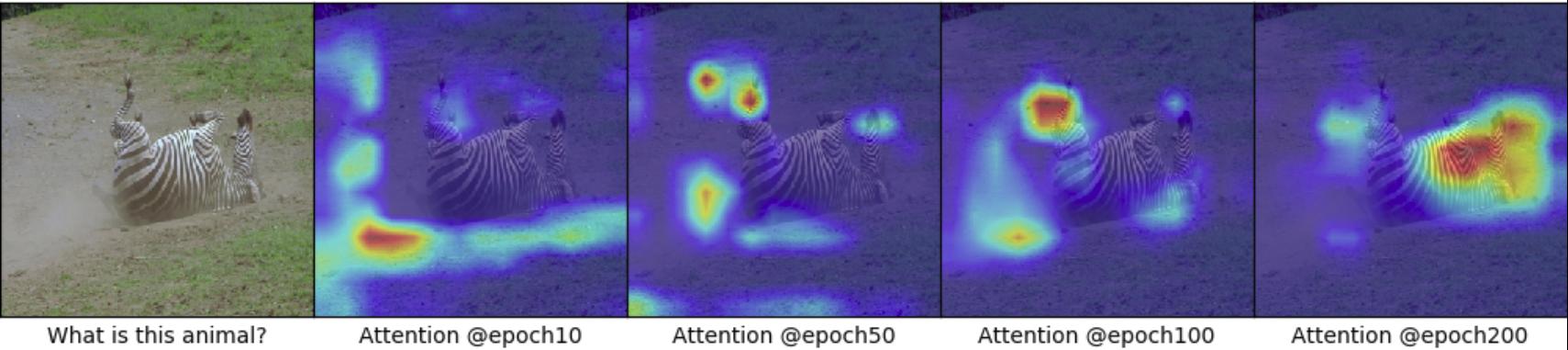}\\
	\vspace{-0.7em}
	\caption{Visualisation of attention map after epoch-10, epoch-50, epoch-100, epoch-200.}
	\label{fig:training}
	\vspace{-1.4em}
\end{figure}
\subsection{Why adversarial learning rather than supervised learning?}
Attention and Grad-CAM distributions differ as has been pointed out. However, the Grad-CAM results are based on the true labels. Therefore, if the distributions are close, then it would serve the purpose. This is because, the attention maps need not exactly correspond to the gradient of the class activations. By using adversarial learning and trying to fool the discriminator, we are able to serve our purpose. This is ensured also by providing comparisons against explicitly using Grad-CAM as supervision with MSE loss results in lower performance. Therefore, adversarial learning is a good method for solving this problem (better even than other distribution matching techniques such as MMD or CORAL). To validate this, we conduct an experiment on distribution matching between the generated attention mask and the ground truth explanation mask. One of the simplest ways to measure the overlapping distributions is the Wasserstein ~\citep{arjovsky_STAT2017towards} distance between them.  We observe that for a perfect adversarial game, the model achieves pure or Mixed Nash Equilibrium, the joint distribution between p (explanation) distribution and q (attention) distribution should be diagonal, that is p \& q distribution are highly overlapped. And the EMD should be very small.  So, using Grad-CAM supervision for attention mask helps to achieve more close towards Mixed Nash Equilibrium in two player game as compared to random and RISE based supervision.  We also observe that if the overlapping region between p-distribution and q-distribution is very low, then KL-divergence in our adversarial game completely fails and JS-divergence works well. In this experiment, we consider three types of explanations mask, Grad-CAM ~\citep{selvaraju2017grad},  RISE ~\citep{petsiuk_BMVC2018rise} and a random mask. We start to observe that with a random explanation mask the accuracy is not improving; rather it is decreasing, when the overlap of the distribution increases, the performance in terms of rank correlation and accuracy is also increasing. We show the experiment result for $PAAN\_SAN\_Ran\_07$ whose  distribution overlapping is 7\% and $PAAN\_SAN\_Ran\_20$ distribution overlapping is 20\%  as shown in table-\ref{abl_VQA1_accuracy} and second last row of table -\ref{tab_rank_correlation}.

We conduct another type of experiment to analyze the contribution of the adversarial loss function used in equation-~\ref{e4} \& ~\ref{e7}. In this experiment we vary the value of $\eta$ in order of magnitude range from {0, 0.1, 0.01, 1, 10, and 100}. We observe that for low value $\eta$, the gradient of the discriminator vanishes rapidly and generator learns nothing.  For $\eta=10$, we get the best result in-terms of VQA accuracy result and rank correlation.
  \begin{figure*}[!htb]
     \small
     \begin{tabular}{ c  c }
     \includegraphics[width=0.5\textwidth,height=4.75cm]{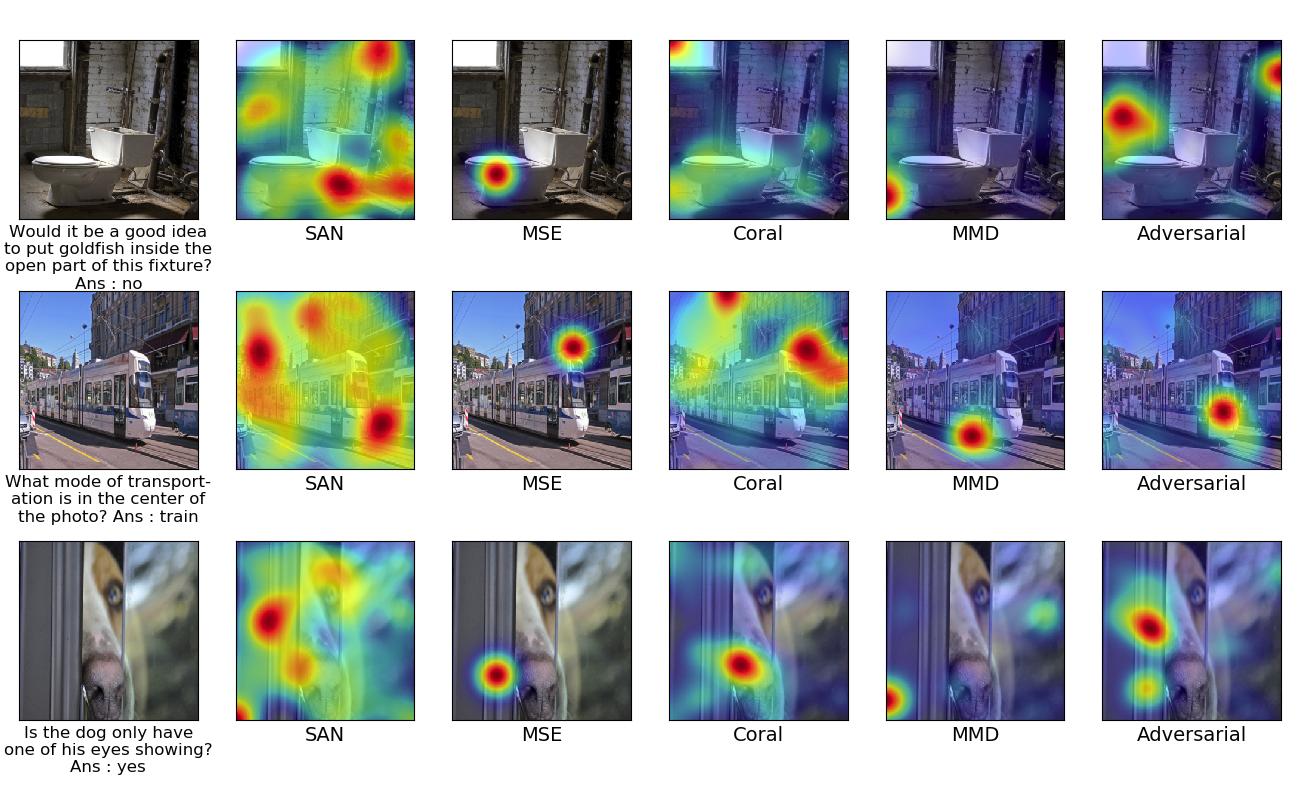}
     & \includegraphics[width=0.5\textwidth,height=4.75cm]{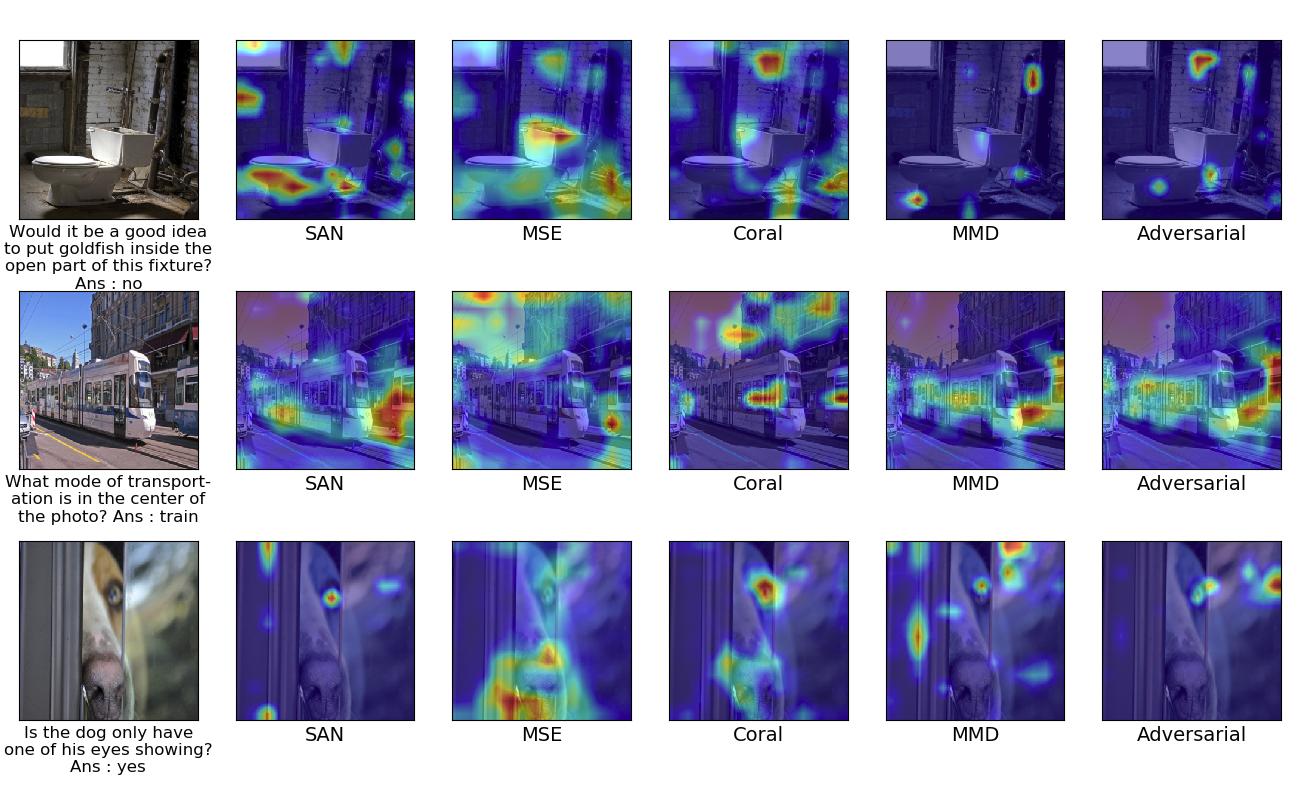}
       \end{tabular}
       \vspace{-1.4em}
      \caption{Examples with different approaches in each column for improving attention using explanation in a self supervised manner. The first column indicates the given target image and its question and answer. Starting from second column, it indicates the Attention map (left) / Grad-CAM map (right) for Stack Attention Network, MSE based approach, Coral based approach, MMD based approach, Adversarial based approach respectively.}
      \label{fig:result_1_B}
       	\vspace{-1.8em}
 \end{figure*}
 


 As Grad-Cam and attention mask may not match at the beginning, it may create instability in the training. To avoid this issue we have used pre-trained SAN and MCB model for warm-start model. We then use an adversarial loss based on the attention and Grad-CAM.  We follow the same training procedure for SAN and MCB models. The figure-~\ref{fig:training} shows attention mask entropy as epoch increases. Initially the entropy is too high and as time goes the entropy decreases. The entropy difference between epoch-10 and epoch-100 is too high as compared to epoch-100  \& epoch 200. 
 

\vspace{-0.6em}
\subsection{Comparison with baseline and state-of-the-art}\label{SOTA}
We obtain the initial comparison with the baselines on the rank correlation on human attention (HAT) dataset \citep{Das_EMNLP2016} that provides human attention while solving for VQA. Between humans the rank correlation is 62.3\%. The comparison of various state-of-the-art methods and baselines are provided in table~\ref{tab_rank_correlation}. We use variant of SAN\cite{Yang_CVPR2016} model as our baseline method. We obtain an improvement of around 3.7\% using AAN network and 6.39\% using PAAN network in terms of rank correlation with human attention. We also compare with the baselines on the answer accuracy  on VQA-v1\citep{VQA} and VQA-v2\citep{Goyal_CVPR2017} dataset as shown in table ~\ref{VQA1_accuracy} and table ~\ref{VQA2_accuracy} respectively. We obtain an improvement of around 5.8\% over the comparable baseline.  Further incorporating MCB improves the results for both AAN and PAAN resulting  in an improvement of 7.1\% over dynamic memory network and 3\% improvement over MCB method on VQA-v1 and 4.2\% on VQA-v2. However, as noted by \citep{Das_EMNLP2016}, using a saliency based method  \citep{Judd_ICCV2009} that is trained on eye tracking data to obtain a measure of where people look in a task independent manner results in more correlation with human attention (0.49). However, this is explicitly trained using human attention and is not task dependent. In our approach, we aim to obtain a method that can simulate human cognitive abilities for solving tasks. The method is not limited to classification alone though all the methods proposed for VQA-1 and VQA-2 datasets follow this. The proposed framework can easily be extended to generative frameworks that generate answers in terms of sentences. We use visual dialog task\citep{Das_ICCV2017} for generative framework, where we visualised improved attention map with respect to generated answer. We observe improvement of overall performance in terms of NDGC values by 1.2\% and MRR values by 0.78\% over the baseline dialog model \citep{Das_ICCV2017}.  
We have provided more results of AAN and PAAN for VQA and Visual dialog, attention map visualization, dataset, and evaluation methods in our project page-~\ref{footnote_1}.

\vspace{-0.5em}
\subsection{Qualitative Result}\label{ResultQual}
We provide attention map visualization for all models as shown in Figure \ref{fig:result_1_B}. We can vividly see how attention is improving as we go from our baseline model (SAN) to the proposed adversarial model (PAAN). For example, in the second row, SAN is not able to focus on any specific portion of the image but as we go towards right, it is able to focus near the bus. Same can be seen for other images also. We have also visualized Grad-CAM maps for the same images to corroborate our hypothesis that Grad-CAM is a better way of visualization of network learning as it can focus on the right portions of the image even in our base line model (SAN). Therefore, it can be used as a tutor to improve attention maps. Our PAAN model helps to learn the attention distribution in an adversarial manner from Grad-CAM distribution as compared to SAN and others. Also, Grad-CAM is simultaneously improved according to our assumption and can also be seen in the Figure \ref{fig:result_1_B}. For example, in SAN the focus of Grad-CAM is spread over the image. In our proposed model, visualization is improved to focus only on the required portion. In the project website\footnote{\url{https://delta-lab-iitk.github.io/TwoPlayer/} \label{footnote_1}}, we show variance in attention map for the same question to the image and its composite image in VQA2.0 dataset. 
\vspace{-0.5em}
\section{Conclusion}
 \vspace{-0.5em}
In this paper we have proposed a method to obtain surrogate supervision for obtaining improved attention using visual explanation. Specifically, we consider the use of Grad-CAM. However, other such modules could also be considered. We show that the use of adversarial method to use the surrogate supervision performs best with the pixel-wise adversarial method (PAAN) performing better against other methods of using this supervision. The proposed method shows that the improved attention indeed results in improved results for the semantic task such as VQA or Visual dialog. Our method provides an initial means for obtaining surrogate supervision for attention and in future we would like to further investigate other means for obtaining improved attention.

\small
\bibliographystyle{aaai}
\bibliography{aaai2020.bib}
\end{document}